\documentclass[letterpaper]{article}
\usepackage{iclr2016_conference,times}
\usepackage{helvet}
\usepackage{courier}
\usepackage{hyperref}
\usepackage{graphics}
\usepackage{graphicx}
\usepackage{url}
\usepackage{amsfonts}
\usepackage{array}
\usepackage{color}
\usepackage{paralist}
\usepackage{multirow}
\newcolumntype{P}[1]{>{\centering\arraybackslash}p{#1}}
\usepackage[textsize=tiny]{todonotes}

\graphicspath{{./imgs/}}

\begin{document}
%
\title{Recurrent Reinforcement Learning: \\ A Hybrid Approach}
\author{Xiujun Li$^1$, Lihong Li$^2$, Jianfeng Gao$^2$, Xiaodong He$^2$, Jianshu Chen$^2$, Li Deng$^2$, Ji He$^3$\\
\texttt{lixiujun@cs.wisc.edu} \\
\texttt{\{lihongli, jfgao, xiaohe, jianshuc, deng\}@microsoft.com} \\
\texttt{jvking@uw.edu} \\
$^1$University of Wisconsin - Madison \hspace{5mm}
$^2$Microsoft Research \hspace{5mm}
$^3$University of Washington - Seattle
}
\maketitle

\begin{abstract}
Successful applications of reinforcement learning in real-world problems often require dealing with partially observable states.  It is in general very challenging to construct and infer hidden states as they often depend on the agent's entire interaction history and may require substantial domain knowledge.  In this work, we investigate a deep-learning approach to learning the representation of states in partially observable tasks, with minimal prior knowledge of the domain.  
In particular, we propose a new family of hybrid models that combines the strength of both supervised learning (SL) and reinforcement learning (RL), trained in a joint fashion: The SL component can be a recurrent neural networks (RNN) or its long short-term memory (LSTM) version, which is equipped with the desired property of being able to capture long-term dependency on history, thus providing an effective way of learning the representation of hidden states. The RL component is a deep Q-network (DQN) that learns to optimize the control for maximizing long-term rewards. Extensive experiments in a direct mailing campaign problem demonstrate the effectiveness and advantages of the proposed approach, which performs the best among a set of previous state-of-the-art methods.
\end{abstract}


\newcommand{\Aset}{\mathcal{A}}
\newcommand{\Oset}{\mathcal{O}}
\newcommand{\Rset}{\mathcal{R}}
\newcommand{\R}{\mathbb{R}}
\newcommand{\E}{\mathbb{E}}
\newcommand{\defeq}{:=}
\newcommand{\1}[1]{\mathbf{1}{\{#1\}}}
\newcommand{\hs}{\tilde{h}}
\newcommand{\same}{\textsc{same}}
\newcommand{\diff}{\textsc{deviated}}


\newcommand{\secref}[1]{Section~\ref{#1}}
\newcommand{\figref}[1]{Figure~\ref{#1}}
\newcommand{\tblref}[1]{Table~\ref{#1}}
\newcommand{\namecite}[1]{\citeauthor{#1}~(\citeyear{#1})}

\newcommand{\cf}{\textit{c.f.}}
\newcommand{\ie}{\textit{i.e.}}
\newcommand{\eg}{\textit{e.g.}}
\newcommand{\etc}{\textit{etc.}}
\newcommand{\aka}{\textit{a.k.a.}}
\newcommand{\etal}{\textit{et~al.}}

\newcommand{\todoc}[2][]{\todo[color=green!20!white!80, #1]{#2}}

\newcommand{\lihong}[1]{\todoc{#1}}
\newcommand{\jianfeng}[1]{\textcolor{blue}{\textbf{JF:} #1}}

\renewcommand{\subsubsection}[1]{\paragraph{#1}}

\section{Introduction}
\label{sec:intro}

Consider customer relationship management (CRM) of a firm that interacts with users over time.  At each decision point, the firm takes an action on its users, such as sending a catalog, a coupon or a greeting card.  In response, a user may visit the store, place an order, or simply ignore the action.  The goal of the firm is to take optimal actions to maximize total profits from users.  In marketing, it is well established that actions taken by the firm can have a long-term effect on user response in the future, implying that \emph{myopic} optimization of profit is usually sub-optimal.  Instead, the life-time value (LTV) of users is a more desired metric of interest~\citep{Dwyer97Customer}. With LTV as the objective, CRM can be naturally formulated as a reinforcement-learning (RL) problem~\citep{Sutton98Reinforcement} where the immediate profit is used as a \emph{reward} and LTV as a long-term \emph{value function}. A similar motivation was used in a recent application of RL to advertising~\citep{Theocharous15Personalized}.

Like many other real-world problems, \eg, robotics and human-computer interaction applications, CRM is challenging partly because of the partial observability of a user's \emph{(Markovian) state}.  Roughly speaking, a user's state summarizes her entire interaction history with the firm: conditioned on the state and future actions, future response of the user is independent of the interaction history.  In practice, constructing and measuring such a state is difficult in complex problems like CRM.  Popular choices such as the Recency-Frequency-Monetary value model (details of which are given in experiments) arguably capture only partial information of a real user state.  The problem of state inference therefore becomes critical when applying RL to these non-Markovian problems.

The most common approach to dealing with partially observable states in reinforcement learning is to use a partially observable Markov decision process, or POMDP~\citep{Kaelbling98Planning}, which is found successful in a few domains~\citep{Pineau03Point,Williams07Partially}.  However, defining hidden states in a POMDP requires substantial domain knowledge, while such knowledge is not always available (or hard to obtain) for many complex, real-world tasks.

In this work, inspired by the recent success of deep reinforcement learning~\citep{Mnih15Human}, we investigate the use of deep neural networks to capture and infer hidden states in an automatic way.  As opposed to POMDP-based approaches, deep learning holds the promise of automatically finding appropriate representations for a given problem, which can be difficult for a human expert, thus avoiding the laborious and challenging step of designing hidden states; see \namecite{Deng14Deep} for an extensive survey of successful applications. In this paper, we propose a new, hybrid approach to using deep learning to tackle complex tasks. Our approach differs from previous ones in two aspects:

\begin{itemize}
\item{First, unlike \namecite{Mnih15Human}, we employ recurrent neural networks (RNN) and long short-term memory (LSTM)~\citep{Hochreiter97Long} models to learn the representation of states for RL. Since these recurrent models can aggregate partial information in the past, and can capture long-term dependencies in the sequential information, their performance is expected be superior to the contextual-window-based approach, which was used in the DQN model of \namecite{Mnih15Human}.}

%

\item{Second, in order to best leverage supervision signals in the training data, the proposed hybrid approach combines the strength of both supervised learning and RL. 
In particular, the model in our hybrid approach is jointly learned using stochastic gradient descent (SGD): in each iteration, the representation of hidden states is first inferred using supervision signals (i.e. next observation and reward) in the training data;
then, the Q-function is updated
using the DQN that takes the learned hidden states as input. The superiority of the hybrid approach is validated in extensive experiments on a benchmark dataset.}
\end{itemize}

In the rest of the paper, we will first review background information and related work. Then, 
we describe in detail our new, hybrid approach. The proposed approach is compared with previous methods in a public CRM benchmark in a series of experiments.  Finally, the paper concludes with a discussion of future directions.

\section{Background and Related Work}
\label{sec:prelim}

\subsection{Reinforcement Learning}
\label{sec:prelim-rl}

In reinforcement learning, an agent uses observation and rewards to learn a (near-)optimal policy for an environment that maximizes the expected total reward. Formally, in discrete steps $t=1,2,3,\ldots$, the agent receives an observation $o_t\in\Oset$, takes an action $a_t\in\Aset$, and receives a real-valued reward $r_t$, where $\Oset$ and $\Aset$ are the sets of observations and actions, respectively.  Let $h_t=(o_1,a_1,r_1,\ldots,o_{t-1},a_{t-1},r_{t-1},o_t)$ be an \emph{interaction history} up to step $t$. The agent may select actions according to a \emph{policy} $\pi(h_t)$ at step $t$. The goal of RL is find $\pi$ to maximize
the expected discounted cumulative reward, $R=\E[\sum_t \gamma^{t-1}r_t]$, for a given discount factor $\gamma\in(0,1)$.

In the case of MDPs where observations are states, $o_t$ is often denoted as $s_t$.  The Q-function, $Q^(s,a)$, is the expected discounted cumulative reward obtained by taking action $a$ in state $s$ and then following an optimal policy thereafter.  The celebrated Q-learning algorithm and its variants~\citep{Sutton98Reinforcement} can be used to learn the Q-function from data, by repeated applications of a stochastic approximation update rule on observed transitions $(s,a,r,s')$:
\[
Q(s,a) \leftarrow Q(s,a) + \eta (r + \gamma \arg\max_{a'}Q(s',a') - Q(s,a))\,,
\]
where $\eta$ is a step-size parameter.  Once $Q \approx Q^*$, the greedy policy, $\pi_Q(s) \defeq \arg\max_a Q(s,a)$, is near-optimal.

In non-Markovian problems like POMDPs~\citep{Kaelbling98Planning}, $o_t$ provides partial information about the unobserved state $s_t$, and can be used to sequentially update the \emph{belief state}.  Although POMDPs have a solid theoretical foundation, their application often requires substantial domain knowledge to define the set of hidden states and observation probabilities.
%
%
In this paper, we use the rich family of RNN/LSTM models to represent and learn hidden states in CRM-like tasks, motivated by their excellent capabilities of representation learning without much human intervention.
Finally, another promising approach to modeling non-Markovian problems is to use a predictive state representation, or PSR~\citep{Littman02Predictive}.  While PSR has great representational power and may be easier to learn from data than POMDPs, like POMDPs, applying PSR often requires substantial domain knowledge to design features or a kernel function~\citep{Boots11Closing}.


\subsection{Deep (Reinforcement) Learning}
\label{sec:prelim-drl}

Recently, deep learning has seen exciting successes in solving reinforcement-learning problems.  Most prominent is the recent use of a deep Q-network (DQN) in Q-learning to solve a large number of Atari games~\citep{Mnih15Human}, although neural networks have been used in some of the classic RL applications like TD-Gammon~\citep{Tesauro95Temporal}.

For partially observable environments, deep learning may also be used to represent and track hidden states, without much domain knowledge.  For example, \namecite{Deng13DeepAtMicrosoft} apply a deep network to track belief states in a spoken dialogue system.  Examples in RL include earlier applications of recurrent neural networks to control problems~\citep{Bakker02Reinforcement,Lin93Reinforcement}, and more recently to Atari games~\citep{Hausknecht15deepRecurrentQ}
and text games~\citep{Narasimhan15TextGames}.
%
In these works, an RNN or LSTM is used to represent a Q-function, $Q(s,a;\theta)$, parameterized by $\theta$.  We call these models RL-RNN and RL-LSTM, respectively.  A variant of Q-learning updates parameters on the observed transitions $(s,a,r,s')$ by:
\[
\theta \leftarrow \theta + \eta \left(r + \gamma \max_{a'} Q(s',a') - Q(s,a)\right) \nabla_\theta Q(s,a;\theta)\,.
\]

In contrast to previous work, 
we propose a new, hybrid model which combines supervised learning and reinforcement learning.  During training, we use the supervised signals to learn the state representation, then jointly train DQN to approximate the Q-function.



\subsection{Customer Relationship Management}
\label{sec:prelim-crm}

In general, CRM refers to data-driven approaches to determining corporate practices in order to maximize lifetime value of customers~\citep{Kumar12Customer}.  Central to CRM is the notion of lifetime value~\citep{Dwyer97Customer}, as opposed to short-term measures of customer value.  In the past, (un)supervised learning has been applied to CRM on problems like customer segmentation, although the focus is on obtaining useful insights to support business decision making~\citep[page~7]{Berry04Data}.  Our work, in contrast, tries to close the decision-making loop: we aim to develop machine-learned models that \emph{directly suggest actions to maximize LTV of customers}.  

We thus take an RL approach to learn a decision-making policy from data.  \namecite{Pednault02Sequential} consider cost-effective decision making in CRM, using \emph{batch} Q-learning to learn a piecewise linear Q-function.  Later, \namecite{Silver13Concurrent} apply variants of Q-learning to learn a linear Q-function in the CRM task of email campaigns.  In contrast, we use the much more flexible function approximator of neural networks to learn the Q-function, which substantially outperforms a strong baseline that uses linear Q-functions in experiments.  More importantly, it provides an effective way to deal with hidden states that are not considered by these authors.

Closest to this work is a recent application of DQN to CRM~\citep{Tkachenko15Autonomous}, which uses the same benchmark data.  Our work differs in a number of substantial ways.  First, we focus on the challenge of non-Markovian CRM problems, overcoming the suboptimality associated with his use of DQNs.  Second, we employ state-of-the-art deep learning models 
to capture hidden states, and develop a novel, hybrid model combining the strengths of supervised and reinforcement learning.
Finally, we adopt a different evaluation methodology that is more appropriate for RL tasks, while the one in \namecite{Tkachenko15Autonomous} is fundamentally flawed.  Details of these distinctions will be clearer later in the paper.



\section{Model}
\label{sec:models}

Recall that the our goal is to learn the optimal Q-function from a sequence of (or sequences of) interaction histories in the form of $(o_1,a_1,r_1,o_2,a_2,r_2,\ldots)$.  A common approach is to optimize a recurrent network to approximate the Q-function, given such networks' strong ability to capture long-term dependency.  Once a good approximation is obtained, a near-optimal policy can be readily defined that selects actions greedily.  Such an approach, however, mingles policy learning and long-term dependence learning during network optimization, which makes it challenging to find and train a single network for both purposes simultaneously.

Motivated by this challenge, we propose to use a \emph{hybrid} model with two networks, and more importantly, a novel \emph{joint} optimization procedure for this model.  The model is designed based on the following observations.  reinforced learning (RL) models, as described in ~\ref{sec:rl}, can be trained to maximize long-term rewards.  In contrast, supervised learning (SL) models, as described in ~\ref{sec:sl}, can be optimized to predict observations and immediate rewards, thus having the potential to better represent and infer hidden states.
With such complementary strengths, it is beneficial to take a hybrid approach, which uses SL for hidden-state representation learning and RL for policy learning.  Moreover, these two components should \emph{not} be optimized \emph{separately}: ideally, the SL component should learn an internal state representation that allows the RL component to maximize long-term reward.  Decoupling the training of two networks likely results in a worse learned policy.

\begin{figure}[h]
    \centering
    \includegraphics[width=0.9\columnwidth]{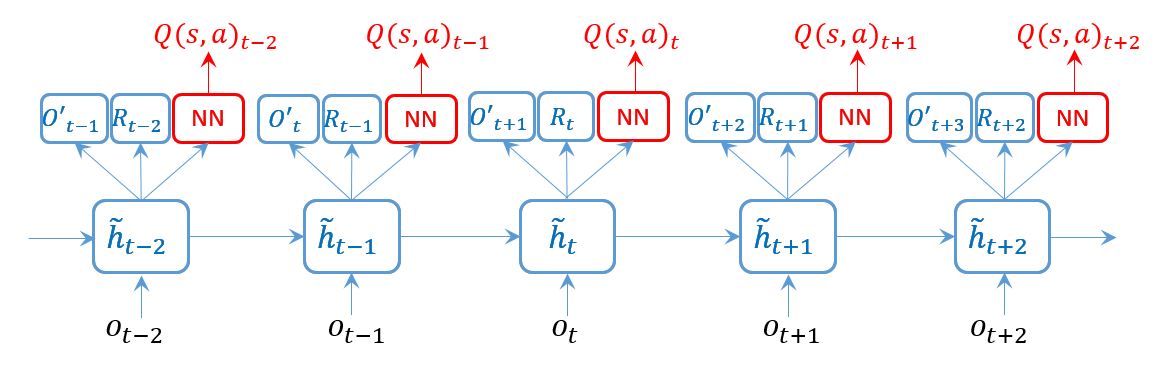}
    \caption{Supervised RNN + Reinforced DQN: $o_t$ is the observation, $\tilde{h}_t$ is the hidden state for RNN, $o_{t+1}'$ is the predicted observation for time $t+1$, $R_t$ is the predicted reward, $Q(s, a)_t$ is the predicted Q-value at time $t$. The blue parts correspond to an unfolded RNN for SL, and the red parts for DQN.  In this hybrid model: the input of DQN is the hidden layers of the supervised RNN model.}
   \label{fig:sl_rnn_rl_dqn}
\end{figure}

Specifically, we propose a new family of \emph{hybrid models} combining supervised learning and reinforcement learning, trained in a \emph{joint} fashion: the SL component can be for example an RNN or LSTM; the RL component is a DQN.  The resulting models are called SL-RNN+RL-DQN (\figref{fig:sl_rnn_rl_dqn}) and SL-LSTD+RL-DQN, respectively.

For training the hybrid model, we use a  joint supervised-reinforced approach.  First, we train an RNN (or LSTM), which learns hidden states from signals including next observations and immediate rewards.  Then, the learned hidden states are the input to DQN, which learns Q-function of a near-optimal policy.  These two training steps are interleaved in each SGD iteration.


The difference between these models and RL-RNN/RL-LSTM~\citep{Bakker02Reinforcement,Hausknecht15deepRecurrentQ,Lin93Reinforcement,Narasimhan15TextGames}. is that, during training, the supervised signals are used to learn the state information, and are back-propagated to the head of RNN/LSTM, while the RL signals are only back-propagated to the hidden layers of RNN for DQN training, and do not involve in the RNN training.

\section{Experiments}
\label{sec:experiments}

\subsection{DataSet}
\label{sec:dataset}

In this paper, we use the 1998 KDD Cup direct mailing dataset\footnote{https://kdd.ics.uci.edu/databases/kddcup98/kddcup98.html}, which has been used in the RL literature~\citep{Marivate15Improved,Pednault02Sequential} for various purposes.  It was collected by a non-profit organization, PVA, who provides programs
and services for US veterans with spinal cord injuries or disease. PVA raises money via direct mailing campaigns. The dataset contains a record for every donor who received the mailing and did not make a
donation in the $12$ months before that. For each of them, it is recorded whether and how much they donated as a response to the campaigns. Apart from that, data is given about the previous and current mailing campaign,
as well as personal information and the giving history of each lapsed donor. The training data is collected for $23$ distinct periods for $95,412$ donors, resulting in over $2$M transition tuples.  Each donor's interaction history can be viewed as a time series of $23$ steps.

Direct mailing campaigns are a typical CRM task, where the decision is on what type of email to send, in order to maximize long-term profit (or cumulative donation in the case of PVA).
In the dataset, we found $12$ actions, including $11$ mailing types and $1$ inaction (corresponding to non-response in the dataset).  The resulting data for each client is a sequence $(o_1,a_1,r_1,\ldots,o_{22},a_{22},r_{22},o_{23})$, where:
\begin{compactitem}
\item{$o_t$ is the current observation of the client.  As in previous work~\cite{Tkachenko15Autonomous}, it is a $5$-dimensional vector consisting of (1) how recently the donor donated last, (2) how frequently she donates, (3) her average donation amount, (4) how many times PVA sends her a mail in the last six months, and (5) how many times PVA has sent her mails.  The first three correspond to the well-known Recency-Frequency-Monetary value model in CRM, and the other two are application specific.}
\item{$a_t$ is one of the $12$ actions taken by PVA.}
\item{$r_t$ is the immediate reward received after taken action $a_t$. In the raw dataset, the reward is the amount of donation in dollars, ranges from \$$0$ to \$$1000$.
}
%
%
\end{compactitem}

\subsection{Evaluation Methodology}
\label{sec:eval}

Reliable evaluation has been a challenge in reinforcement learning when no simulator (like those used in Atari games) is available.  The approach of \namecite{Tkachenko15Autonomous} proceeds as follows.  After a model is optimized using training data, it is run on test data to select actions in every step.  The test data is then partitioned into two subsets: the \same\ set consists of transitions where the model's selected action is the same as the action in the data; all other transitions are in the \diff\ set.  Clearly, the partition into \same\ and \diff\ is model dependent.  Finally, a model is considered better if its average reward in the corresponding \same\ set is higher.

While the procedure above might sound intuitive and appealing, it is fundamentally flawed for evaluating action-section models.  First, it focuses on myopic rewards, thus fails to reflect the capability of RL models whose very aim is to optimize (long-term) LTV of customers.  The second problem is more subtle but equally severe: that evaluation is really about finding correlation 
rather than causation,
which can lead to many paradoxes.
For instance, imagine a model that learns how PVA selected actions when collecting this dataset.  It can then select actions based on whether the client is generous (using information encoded in observations): it selects the same action as PVA if and only if the client is generous.  This way, the model is able to ``game'' the evaluation protocol of \namecite{Tkachenko15Autonomous} by enforcing its \same\ set to contain only generous clients who tend to donate more.  However, such a cherry-picking model is not expected to do anything better than the data-collection policy of PVA.

Given these important drawbacks, we adopt a different evaluation that is common in the RL literature: we use the dataset to build a simulator, and rely on the simulator to generate synthetic CRM interaction sequences for training and evaluating different action-selection models.  While building a model is in generally nontrivial, this data presents a factored structure: at any step $t$, given $a_t$ and $r_t$, the five components in the observation vectors evolve (from $o_t$ to $o_{t+1}$) independently.  A similar approach was also taken in previous work~\citep{Pednault02Sequential}. 
%
%
%
More specifically, at step $t$, the simulator takes the observed history $h_t$ and action $a_t$ as input to predict:
\begin{compactitem}
\item{next observation $o_{t+1}$: the $5$-dimensional observation is discrete, and individual dimensions evolve independently of each other.  We therefore build an observation probability table for each observation dimension, and then sample next observations using these tables.}
%
\item{reward $r_t$: in the experiments, we build reward function using an RNN,
trained to predict reward $r$ using its internal history summary $\tilde{h}_t$. This simulator thus creates a realistic scenario that allows hidden states and long-term effects on customers~\citep{Netzer08Hidden}.}
\end{compactitem}

\subsection{Experiment Setup}
\label{sec:setup}

We found smaller data are enough to yield strong policies, therefore only use a random subset of donors of the entire data for experiments.  We tried four data sizes of varying number of donors, each having $23$ steps, so that the total number of transitions is \{$50$K, $100$K, $200$K, $500$K\}.  The data were then split into training, validation, and test sets with proportions 4:1:1.

%
%
%
To generate training data, we started with the initial observation vector of donors in the training set, and ran one of the following data-collection policies to select actions:
\begin{itemize}
\item{Uniformly random (U): at any step, an action is chosen uniformly at random from the set of actions.}
\item{Probability-matching (M): at any step, an action is chosen with probability proportional to its frequency in raw data.}
%
\item{Real (R): the chosen action is the actual one recorded in raw data.  The simulator is used only to regenerate reward.}
\end{itemize}



\tblref{tab:eva_setting} summarizes all choices along three dimensions in our experiment setup.  We fixed $\gamma=0.9$ and compared the average per-step reward (donation) of the learned policies. We ran each setting at least $5$ times and report the average.

The following models are used as baselines in our experiments:
\begin{itemize}
\item{One approach is to treat the problem as supervised learning (SL), where we predict whether the short-term reward $r_t$ is larger than a threshold $\tau$.  The input can be the current observation $o_t$ or the sequence of observations $(o_1,\ldots,o_t)$, leading to a multi-layer neural networks (DNN) and RNN/LSTM, respectively.  In our experiments, we found $\tau=0$ to work best empirically.  At test time, the model, denoted $\hat{R}$, takes current observation or the observation sequence as input, and selects actions greedily according to its reward predictions.}
\item{The DQN~\citep{Mnih15Human} with history window is a deep RL baseline.  We select the best model from candidate history window lengths $\{1,2,3\}$, and report the best performance.}
\item{Two other deep RL baselines, RL-RNN and RL-LSTM, are similar to DQN, but are expected to handle partial observability by explicitly modeling long-term dependencies of future rewards on history~\citep{Bakker02Reinforcement,Hausknecht15deepRecurrentQ,Lin93Reinforcement,Narasimhan15TextGames}.  }
\item{Joint models with separate training: we use the same models described in \secref{sec:models}, but training is different.  We first train RNN/LSTM to minimize square prediction error of observation/reward until the network converges.  Then they are fixed and are used to generate a hidden state representation that is the input to DQN.}
\end{itemize}
More details of the baselines are described in Appendix~\ref{sec:baseline}.

\begin{table}[t]
\caption{Summary of Evaluation Settings}
\label{tab:eva_setting}
\begin{center}
\begin{tabular}{c|c}
{\bf Behavior Policy}  & {\bf Data Size}\\
\hline \hline
U, M, R & $50$K, $100$K, $200$K, $500$K\\
\end{tabular}
\end{center}
\end{table}

\subsection{Results}
\label{sec:results}

Three sets of experiments were done, as summarized in \tblref{tab:eva_setting_group}.  Each experiment focuses on a particular aspect and is discussed in detail in the following subsections.



\begin{table}[t]
\caption{Evaluation Setting Configuration}
\begin{center}
\begin{tabular}{l|c|c}
 & {\bf Behavior Policy} & {\bf Data Size}\\
\hline \hline
E1 & M & $100$K \\
E2 & \{U, M, R\} & $100$K \\
E3 & M & \{$50$K, $100$K, $200$K, $500$K\} \\
\end{tabular}
\end{center}

\label{tab:eva_setting_group}
\end{table}

\subsubsection{Experiment E1}

\begin{table}[t]
\caption{Supervised Learning and Reinforcement Learning under RNN Simulator.
The superscripts $a$, $b$, $c$ and $d$ indicate statistically significant improvements (p$<$0.05) over DNN, SL models, \{SL models, DQN\}, \{RL-RNN, RL-LSTM\} respectively. SL-RNN + RL-DQN$^*$, and SL-LSTM + RL-DQN$^*$ are separate training, $e$ indicates the joint training of hybrid models significantly improve over the corresponding model with separated training.}
\begin{center}
\begin{tabular}{cccc}
\multicolumn{1}{c}{SL Models} &\multicolumn{1}{c}{\bf Avg Reward (\$)}  &\multicolumn{1}{c}{\bf RL Models} &\multicolumn{1}{c}{\bf Avg Reward (\$)}\\
\hline \\
\centering{DNN} & {$8.10$} & {DQN} & {$9.14^a$} \\
\centering{RNN} & {$9.03$} & {RL-RNN} & {$9.39^c$} \\
\centering{LSTM} & {$9.01$} & {RL-LSTM} & {$9.35^c$} \\
\centering{} & {} & {SL-RNN + RL-DQN} & {$9.66^d$} \\
\centering{} & {} & {SL-LSTM + RL-DQN} & {$9.61^d$} \\
\centering{} & {} & {SL-RNN + RL-DQN$^*$} & {$9.49$} \\
\centering{} & {} & {SL-LSTM + RL-DQN$^*$} & {$9.37^e$} \\
\end{tabular}
\end{center}
\label{tab:rnn_results}
\end{table}

This experiment is to investigate how hidden states affect relative performance of various models.  Recall that with the RNN simulator, rewards are a function of the current observation \emph{as well as} history up to that step.  In other words, the model must be able to infer and track such hidden states in order to maximize cumulative rewards.

\begin{figure}
\caption{Learning curve for RL models} \label{fig:converge_curve}
\centering
\includegraphics[width=0.8\columnwidth]{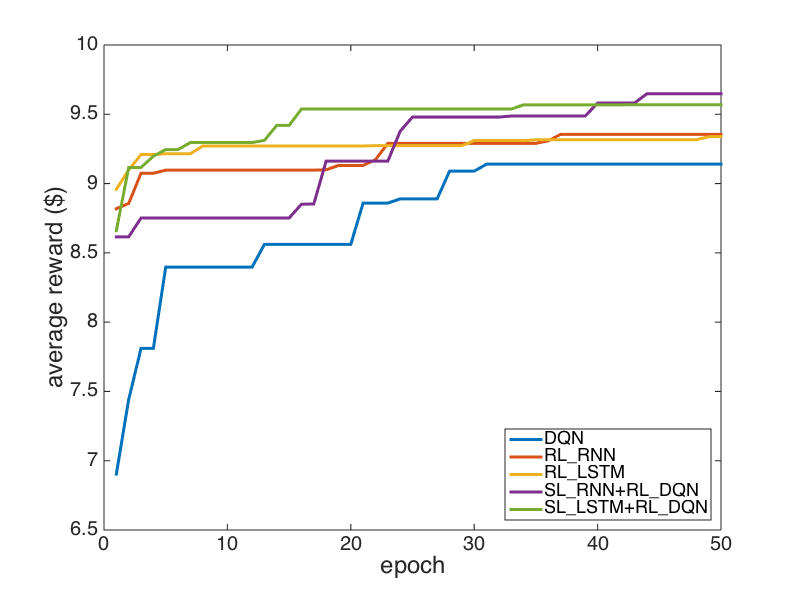}
\end{figure}

From \tblref{tab:rnn_results} and \figref{fig:converge_curve}, we can see that DQN significant outperform DNN. RL-RNN and RL-LSTM significantly outperform all SL models.  Furthermore, there is a clear advantage of RL+RNN and RL+LSTM over DQN, since the latter ignores history information. Third, ``SL-RNN + RL-DQN'' is significantly better than RL-RNN, and similarly when RNN is replaced by LSTM.

Finally, \tblref{tab:rnn_results}, also shows that separated parameter training is inferior to our proposed joint training approach.
With separated training, it is difficult to know whether and when the learned hidden state representation is good enough to enable DQN to learn a good Q-function.  Our joint training approach, on the other hand, couples training of RNN/LSTM and of DQN.  It thus can learn a state representation that facilitates Q-function learning in RL-DQN.

\subsubsection{Experiment E2}

\begin{figure}[h]
    \centering
    \includegraphics[width=1.0\columnwidth]{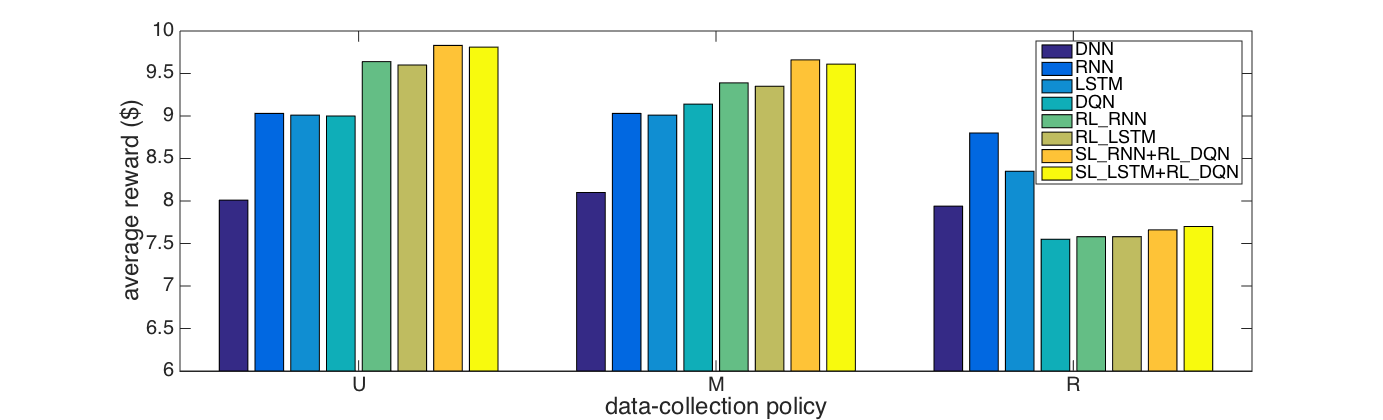}
    \caption{Supervised Learning and Reinforcement Learning under RNN Simulator with different simulation data (U, M, R). Each group has eight models: three SL models and five RL models.}
   \label{fig:rnn_res_umr}
\end{figure}

The second set of experiments is to investigate how the data-collection policy affects model performance.  It is well-known that proper exploration is necessary to learn a good policy, and we examine how it affects our models empirically.  From \figref{fig:rnn_res_umr}, when actions were selected by U and M, qualitatively similar results can be obtained as before.  However, when actions were chosen by R, all models' performance decays, and reinforcement learning models are hurt much more that they are inferior to supervised learning models.

An examination of the raw data revealed that the actual data-collection policy by PVA seemed to be deterministic: the same action is applied to all donors at the same step.  Therefore, little or no exploration exists in this dataset. Reinforcement learning algorithms tend to be more sensitive to lack of exploration than supervised learning algorithms, because the former in a sense try to predict what will happen far into the future.  The less exploration, the more error is introduced into the projection of an RL algorithm, which is consistent with what is shown in 
\figref{fig:rnn_res_umr}.

\subsubsection{Experiment E3}

\begin{figure}[h]
    \centering
    \includegraphics[width=1.0\columnwidth]{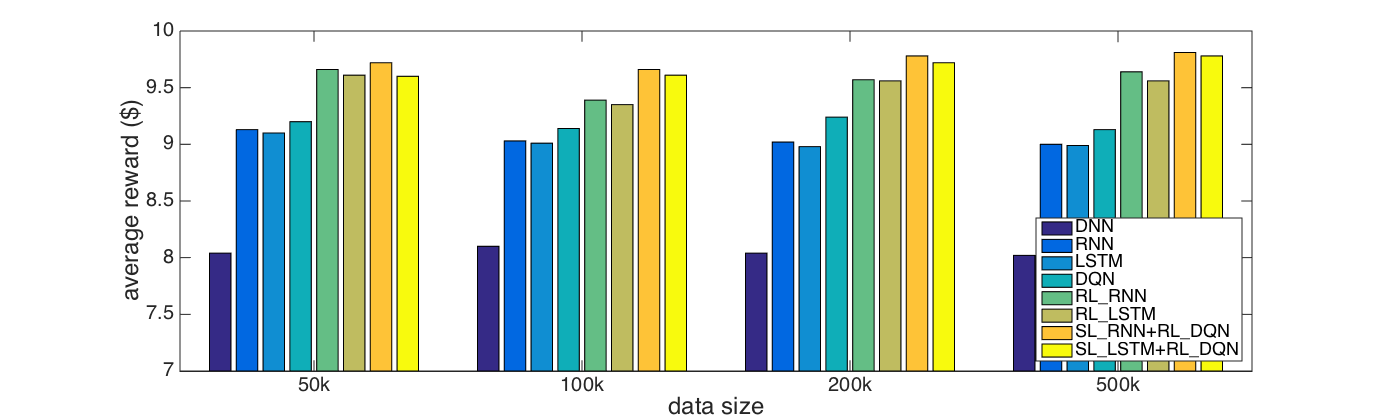}
    \caption{Supervised Learning and Reinforcement Learning under RNN Simulator with different data size ($50K$, $100K$, $200K$, and $500K$). Each group has eight models: three SL models and five RL models.}
   \label{fig:rnn_res_E3}
\end{figure}

In the last set of experiments, we varied data sizes to see how it affects each model's performance.  
From \figref{fig:rnn_res_E3}, 
we saw similar results with a wide range of data sizes.  These preliminary results indicate our models are data-efficient, and the benefits over SL or RL models are consistent with different data sizes. 

\section{Conclusions}
\label{sec:conclusions}

In this work, we 
propose a hybrid approach that uses recurrent deep learning models, and combines the strength of both supervised learning and reinforcement learning, to solve a CRM task, which is typical of real-world non-Markovian problems.  In particular, our hybrid approach utilizes supervised signals in training data to learn hidden-state representations, and then jointly trains an DQN (using reinforcement learning) to optimize the control for maximizing long-term rewards. Through a large-scale experimental analysis under different settings, we showed that the proposed hybrid models significantly outperform other state-of-the-art SL/RL models across the board: (1) Deep RL is more effective than SL for optimizing lifetime values; (2) RL with RNN/LSTM models is a promising approach to solving non-Markovian tasks with long-term dependencies; (3) It is promising to use memory networks models to learn hidden-state representations in a supervised learning manner, with the DQN jointly trained for non-Markovian tasks.  Beyond the analysis, our experimental results demonstrate the promise of deep reinforcement learning for specific CRM tasks. 



The experimental results reported in this paper suggest multiple interesting directions for future work.  One is to explore the use of recurrent networks in model-based or policy-based RL, as opposed to the value-function-based approaches taken in this work.  
Another important direction is to capture latent structures of actions, 
in order to facilitate generalization across actions and to handle newly emerged actions common in a range of applications.


\subsection*{Acknowledgments}

We thank Nan Jiang for helpful discussions on building the simulator in our experiments.

\bibliographystyle{iclr2016_conference}
\bibliography{refs}

\begin{thebibliography}{24}
\providecommand{\natexlab}[1]{#1}
\providecommand{\url}[1]{\texttt{#1}}
\expandafter\ifx\csname urlstyle\endcsname\relax
  \providecommand{\doi}[1]{doi: #1}\else
  \providecommand{\doi}{doi: \begingroup \urlstyle{rm}\Url}\fi

\bibitem[Bakker(2002)]{Bakker02Reinforcement}
Bakker, Bram.
\newblock Reinforcement learning with long short-term memory.
\newblock In \emph{Advances in Neural Information Processing Systems 14
  (NIPS-01)}, pp.\  1475--1482, 2002.

\bibitem[Berry \& Linoff(2004)Berry and Linoff]{Berry04Data}
Berry, Michael~J.A. and Linoff, Gordon~S.
\newblock \emph{Data Mining Techniques: For Marketing, Sales, and Customer
  Relationship Management}.
\newblock John Wiley \& Sons, 2nd edition, 2004.
\newblock ISBN 978-0-471-47064-9.

\bibitem[Boots et~al.(2011)Boots, Siddiqi, and Gordon]{Boots11Closing}
Boots, Byron, Siddiqi, Sajid~M., and Gordon, Geoffrey~J.
\newblock Closing the learning-planning loop with predictive state
  representations.
\newblock \emph{The International Journal of Robotics Research}, 30\penalty0
  (7):\penalty0 954--966, 2011.

\bibitem[Deng \& Yu(2014)Deng and Yu]{Deng14Deep}
Deng, Li and Yu, Dong.
\newblock Deep learning: Methods and applications.
\newblock \emph{Foundations and Trends in Signal Processing}, 7\penalty0
  (3-4):\penalty0 197--387, 2014.

\bibitem[Deng et~al.(2013)Deng, Li, Huang, Yao, Yu, Seide, Seltzer, Zweig, He,
  Williams, Gong, and Acero]{Deng13DeepAtMicrosoft}
Deng, Li, Li, Jinyu, Huang, Jui-Ting, Yao, Kaisheng, Yu, Dong, Seide, Frank,
  Seltzer, Michael, Zweig, Geoff, He, Xiaodong, Williams, Jason, Gong, Yifan,
  and Acero, Alex.
\newblock Recent advances in deep learning for speech research at {Microsoft}.
\newblock In \emph{Acoustics, Speech and Signal Processing (ICASSP-13)}, 2013.

\bibitem[Dwyer(1997)]{Dwyer97Customer}
Dwyer, F.~Robert.
\newblock Customer lifetime valuation to support marketing decision making.
\newblock \emph{Journal of Direct Marketing}, 11\penalty0 (4):\penalty0 6--13,
  1997.

\bibitem[Hausknecht \& Stone(2015)Hausknecht and
  Stone]{Hausknecht15deepRecurrentQ}
Hausknecht, Matthew and Stone, Peter.
\newblock Deep recurrent {Q}-learning for partially observable {MDPs}, 2015.
\newblock arXiv preprint arXiv:1507.06527.

\bibitem[Hochreiter \& Schmidhuber(1997)Hochreiter and
  Schmidhuber]{Hochreiter97Long}
Hochreiter, Sepp and Schmidhuber, J\"urgen.
\newblock Long short-term memory.
\newblock \emph{Neural Computation}, 9\penalty0 (8):\penalty0 1735--1780, 1997.

\bibitem[Kaelbling et~al.(1998)Kaelbling, Littman, and
  Cassandra]{Kaelbling98Planning}
Kaelbling, Leslie~Pack, Littman, Michael~L., and Cassandra, Anthony~R.
\newblock Planning and acting in partially observable stochastic domains.
\newblock \emph{Artificial Intelligence}, 101\penalty0 (1--2):\penalty0
  99--134, 1998.

\bibitem[Kumar \& Reinartz(2012)Kumar and Reinartz]{Kumar12Customer}
Kumar, V. and Reinartz, Werner.
\newblock \emph{Customer Relationship Management: Concept, Strategy, and
  Tools}.
\newblock Springer Texts in Business and Economics. Springer, 2012.
\newblock ISBN 978-3-642-20109-7.

\bibitem[Lin(1993)]{Lin93Reinforcement}
Lin, Long-Ji.
\newblock \emph{Reinforcement Learning for Robots using Neural Networks}.
\newblock PhD thesis, School of Computer Science, Carnegie Mellon University,
  Pittsburgh, PA, 1993.

\bibitem[Littman et~al.(2002)Littman, Sutton, and Singh]{Littman02Predictive}
Littman, Michael~L., Sutton, Richard~S., and Singh, Satinder~P.
\newblock Predictive representations of state.
\newblock In \emph{Advances in Neural Information Processing Systems 14
  (NIPS-01)}, pp.\  1555--1561, 2002.

\bibitem[Marivate(2015)]{Marivate15Improved}
Marivate, Vukosi~N.
\newblock \emph{Improved Empirical Methods in Reinforcement-Learning
  Evaluation}.
\newblock PhD thesis, Rutgers University, New Brunswick, NJ, 2015.

\bibitem[Mnih et~al.(2015)Mnih, Kavukcuoglu, Silver, Rusu, Veness, Bellemare,
  Graves, Riedmiller, Fidjeland, Ostrovski, Petersen, Beattie, Sadik,
  Antonoglou, King, Kumaran, Wierstra, Legg, and Hassabis]{Mnih15Human}
Mnih, Volodymyr, Kavukcuoglu, Koray, Silver, David, Rusu, Andrei~A., Veness,
  Joel, Bellemare, Marc~G., Graves, Alex, Riedmiller, Martin, Fidjeland,
  Andreas~K., Ostrovski, Georg, Petersen, Stig, Beattie, Charles, Sadik, Amir,
  Antonoglou, Ioannis, King, Helen, Kumaran, Dharshan, Wierstra, Daan, Legg,
  Shane, and Hassabis, Demis.
\newblock Human-level control through deep reinforcement learning.
\newblock \emph{Nature}, 518\penalty0 (7540):\penalty0 529--533, 2015.

\bibitem[Narasimhan et~al.(2015)Narasimhan, Kulkarni, and
  Barzilay]{Narasimhan15TextGames}
Narasimhan, Karthik, Kulkarni, Tejas, and Barzilay, Regina.
\newblock Language understanding for text-based games using deep reinforcement
  learning.
\newblock In \emph{Proceedings of the Conference on Empirical Methods in
  Natural Language Processing (EMNLP-15)}, 2015.

\bibitem[Netzer et~al.(2008)Netzer, Lattin, and Srinivasan]{Netzer08Hidden}
Netzer, Oded, Lattin, James~M., and Srinivasan, V.
\newblock A hidden {Markov} model of customer relationship dynamics.
\newblock \emph{Marketing Science}, 27\penalty0 (2):\penalty0 185--204, 2008.

\bibitem[Pednault et~al.(2002)Pednault, Abe, and
  Zadrozny]{Pednault02Sequential}
Pednault, Edwin P.~D., Abe, Naoki, and Zadrozny, Bianca.
\newblock Sequential cost-sensitive decision-making with reinforcement
  learning.
\newblock In \emph{Proceedings of the Eighth International Conference on
  Knowledge Discovery and Data Mining (KDD-02)}, pp.\  259--268, 2002.

\bibitem[Pineau et~al.(2003)Pineau, Gordon, and Thrun]{Pineau03Point}
Pineau, Joelle, Gordon, Geoffrey~J., and Thrun, Sebastian.
\newblock Point-based value iteration: An anytime algorithm for {POMDPs}.
\newblock In \emph{Proceedings of the Eighteenth International Joint
  Conferences on Artificial Intelligence (IJCAI-03)}, pp.\  1025--1032, 2003.

\bibitem[Silver et~al.(2013)Silver, Newnham, Barker, Weller, and
  McFall]{Silver13Concurrent}
Silver, David, Newnham, Leonard, Barker, David, Weller, Suzanne, and McFall,
  Jason.
\newblock Concurrent reinforcement learning from customer interactions.
\newblock In \emph{Proceedings of the Thirtieth International Conference on
  Machine Learning (ICML-13)}, pp.\  924--932, 2013.

\bibitem[Sutton \& Barto(1998)Sutton and Barto]{Sutton98Reinforcement}
Sutton, Richard~S. and Barto, Andrew~G.
\newblock \emph{Reinforcement Learning: An Introduction}.
\newblock MIT Press, Cambridge, MA, March 1998.
\newblock ISBN 0-262-19398-1.

\bibitem[Tesauro(1995)]{Tesauro95Temporal}
Tesauro, Gerald.
\newblock Temporal difference learning and {TD-Gammon}.
\newblock \emph{Communications of the ACM}, 38\penalty0 (3):\penalty0 58--68,
  March 1995.

\bibitem[Theocharous et~al.(2015)Theocharous, Thomas, and
  Ghavamzadeh]{Theocharous15Personalized}
Theocharous, Georgios, Thomas, Philip~S., and Ghavamzadeh, Mohammad.
\newblock Personalized ad recommendation systems for life-time value
  optimization with guarantees.
\newblock In \emph{Proceedings of the Twenty-Fourth International Joint
  Conference on Artificial Intelligence (IJCAI-15)}, 2015.

\bibitem[Tkachenko(2015)]{Tkachenko15Autonomous}
Tkachenko, Yegor.
\newblock Autonomous {CRM} control via {CLV} approximation with deep
  reinforcement learning in discrete and continuous action space, 2015.
\newblock arXiv:1504.01840.

\bibitem[Williams \& Young(2007)Williams and Young]{Williams07Partially}
Williams, Jason~D. and Young, Steve~J.
\newblock Partially observable {Markov} decision processes for spoken dialog
  systems.
\newblock \emph{Computer Speech and Language}, 21\penalty0 (2):\penalty0
  393--422, 2007.

\end{thebibliography}

\appendix

\section{Baseline Models}
\label{sec:baseline}

This section gives additional details for the supervised-learning and deep RL baselines we used in the experiments.

\subsection{Supervised Learning}
\label{sec:sl}

\begin{figure}[h]
    \centering
    \includegraphics[width=0.8\columnwidth]{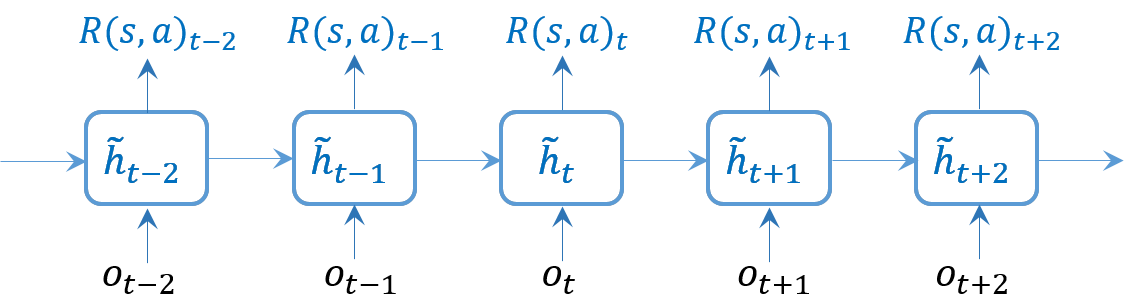}
    \caption{An unfolded supervised learning RNN: $o_t$ is the observation, $\hs_t$ is the hidden state for RNN, $R(s, a)_t$ is the predicted reward at time $t$, where $s$ is the $\hs_t$ of RNN.}
   \label{fig:sl_rnn}
\end{figure}

The first baseline is to treat the problem as SL, in which one tries to predict which action leads to higher expected (immediate) reward given the interaction history so far.  In our experiments, we formulated the problem as regression with the raw reward signal as target.  For each transition tuple $(o,a,r,o')$ from training data, we tried to learn the regression of $r$ given observation $o$ and possibly the history that led to this transition, depending on which network model is used.  This reduction results in standard deep learning models with mean squared error as the loss function for training.
Several network architectures are used, with and without built-in modeling of long-term dependency:
\begin{itemize}
\item{Multi-layer (deep) neural networks (DNN) breaks an interaction history into individual transitions, $\{(o_t, a_t, r_t, o_{t+1})\}_{t=1,2,\ldots}$.  The network is learned to predict $r_t$ based on $(o_t,a_t)$, for $r_t > \tau$.  In our experiments, we found $\tau=0$ to work best empirically.
At test time, the model, denoted $\hat{R}$, takes current observation $o$ as input, and selects actions greedily according to its reward predictions: $\arg\max_a\hat{R}(o,a)$.
}
\item{RNN and LSTM can model long-term dependency in a customer's interaction history.  As shown in \figref{fig:sl_rnn} for the case of RNN, the interaction history can \emph{no longer} be decomposed into separate transitions like DNN.  At step $t$, the model is updated using observation $o_t$, reward $r_t$ and the current internal history summary, $\hs_{t-1}$, which is maintained recursively in RNN.  At test time, the model selects actions in a similar fashion, based on both the current observation and the current internal history summary.  The case for LSTM is similar.}
\end{itemize}

\subsection{Reinforcement Learning}
\label{sec:rl}

SL models above only considers immediate rewards.  In contrast, RL takes future rewards into account and aims to optimize total long-term reward directly, which is desired in CRM task when one tries to optimize LTV of customers.

Our first deep RL baseline is DQN~\cite{Mnih15Human}, where we treat $o_t$ as state $s_t$, 
and optimize network parameters to obtain an approximate Q-function, $Q(s,a)$.  Once a good Q-network is learned, it can be used to select actions in a greedy fashion: $\pi_Q(s) \defeq \arg\max_a Q(s,a)$.

The second and third deep reinforcement learning baselines, RL-RNN and RL-LSTM, are similar to DQN, but are expected to handle partial observability by explicitly modeling long-term dependencies of future rewards on history~\cite{Bakker02Reinforcement,Hausknecht15deepRecurrentQ,Lin93Reinforcement,Narasimhan15TextGames}.  Similar to RNN and LSTM in supervised-learning models, the Q-network now is a function of the current observation $o_t$ and the current internal history summary $\hs_{t-1}$.  Again, the internal history summary $\hs_t$ is updated recursively as time goes on.  Actions are selected greedily after the Q-network is learned.



\begin{figure}[h]
    \centering
    \includegraphics[width=0.8\columnwidth]{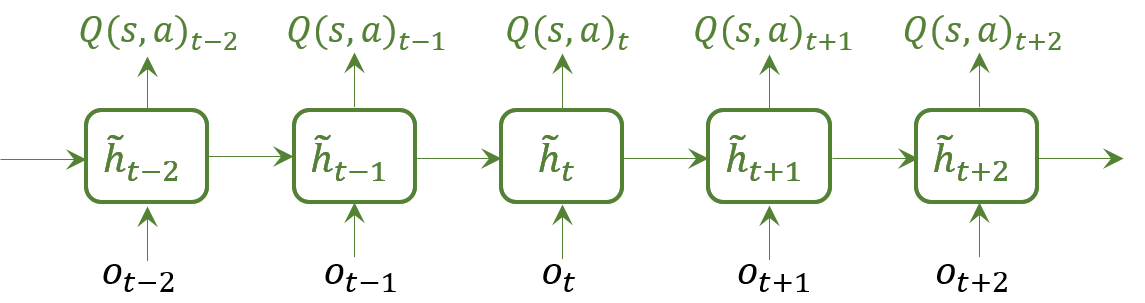}
    \caption{Reinforcement learning with RNN (RL-RNN): $o_t$ is the observation, $\hs_t$ is the hidden state for RNN, $Q(s, a)_t$ is the predicted Q-value for action $a$ at time $t$, and $s$ is $\hs_t$.}
   \label{fig:rl_rnn}
\end{figure}

In practice, training DQN or RL-RNN/RL-LSTM may be unstable, due to dependence of transition tuples in an interaction history. One variant, as used in previous work~\cite{Mnih15Human}, is to use two Q-networks: one network (the ``target network'') is used to define the target value in Q-learning updates, $r+\gamma\max_{a'}Q(s',a';\theta)$, while the other is used for parameter updates.  When the latter network's parameter converges, it becomes the target network, and the process repeats until convergence.  We use the same variant for DQN, RL-RNN and RL-LSTM in the experiments.

\end{document}